\documentclass{article} % For LaTeX2e
\usepackage{iclr2024_conference,times}
\usepackage{mathptmx}
\usepackage{url}
\usepackage{multirow}
\usepackage{siunitx}
\usepackage{booktabs}
\usepackage{fnpos,amssymb,pifont}
\usepackage{graphicx}
\usepackage{natbib}
\usepackage{amsmath,amsfonts,bm}

\def\eqref#1{equation~\ref{#1}}
\def\1{\bm{1}}

\DeclareMathAlphabet{\mathsfit}{\encodingdefault}{\sfdefault}{m}{sl}
\SetMathAlphabet{\mathsfit}{bold}{\encodingdefault}{\sfdefault}{bx}{n}

\usepackage{hyperref}
\usepackage{url}

\title{MEMTrack: A deep learning-based approach to microrobot tracking in dense and low-contrast environments }

\author{%
 Medha Sawhney$^*$ \\
 Dept. of Computer Science\\
  Virginia Tech \\
  medha@vt.edu
  \And
  Bhas Karmarkar$^*$ \\
  Dept. of Mechanical Engg.\\
  Virginia Tech \\
 bhasnk@vt.edu
  \And
  Eric J. Leaman \\
  Dept. of Mechanical Engg.\\
 Virginia Tech \\
  leamanej@vt.edu
  \AND
  Arka Daw \\
  Dept. of Computer Science\\
  Virginia Tech \\
  darka@vt.edu
  \And
  Anuj Karpatne \\
  Dept. of Computer Science\\
  Virginia Tech \\
  karpatne@vt.edu
   \And
  Bahareh Behkam \\
  Dept. of Mechanical Engg.\\
  Virginia Tech \\
 behkam@vt.edu
}
\iclrfinalcopy % Uncomment for camera-ready version, but NOT for submission.
\begin{document}

\maketitle

\begin{abstract}
Tracking microrobots is challenging, considering their minute size and high speed. As the field progresses towards developing microrobots for biomedical applications and conducting mechanistic studies in physiologically relevant media (e.g., collagen), this challenge is exacerbated by the dense surrounding environments with feature size and shape comparable to microrobots. Herein, we report Motion Enhanced Multi-level Tracker (MEMTrack), a robust pipeline for detecting and tracking microrobots using synthetic motion features, deep learning-based object detection, and a modified Simple Online and Real-time Tracking (SORT) algorithm with interpolation for tracking. Our object detection approach combines different models based on the object’s motion pattern. We trained and validated our model using bacterial micro-motors in collagen (tissue phantom) and tested it in collagen and aqueous media. We demonstrate that MEMTrack accurately tracks even the most challenging bacteria missed by skilled human annotators, achieving precision and recall of 77\% and 48\% in collagen and 94\% and 35\% in liquid media, respectively. Moreover, we show that MEMTrack can quantify average bacteria speed with no statistically significant difference from the laboriously-produced manual tracking data. MEMTrack represents a significant contribution to microrobot localization and tracking, and opens the potential for vision-based deep learning approaches to microrobot control in dense and low-contrast settings. All source code for training and testing MEMTrack and reproducing the results of the paper have been made publicly available {\url{https://github.com/sawhney-medha/MEMTrack}}.

\end{abstract}

\section{Introduction}

\begin{figure}[t]
\centering
  \includegraphics[width=0.75\linewidth]{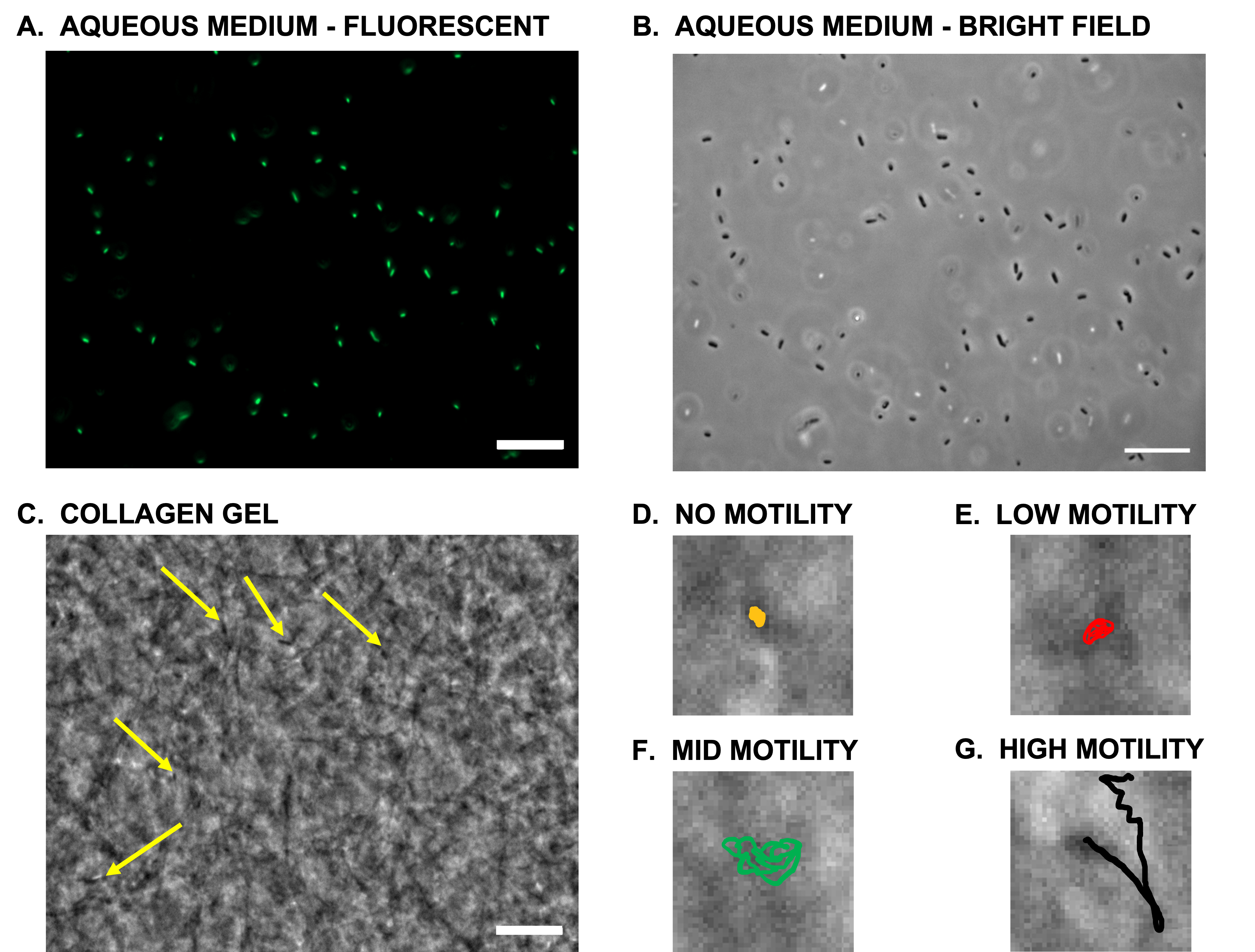}
  \caption{Self-propelled micromotors, such as bacteria, are easy to track in high-contrast fluorescent images (A); however, they are more difficult to detect in high acquisition rate bright-field images in liquid (aqueous) medium (B). The complexity further increases in dense environments such as collagen with feature shapes and sizes similar to those of the microscale objects of interest (C) The yellow arrows represent bacteria locations. (D, E, F, and G) are sample trajectories of bacteria in different motility sub-populations. All scale bars are 20 $\mu$m.}
  \label{fig:motivation}
\end{figure}

In recent years, microrobotic systems have burgeoned due to their potential in various fields, including targeted drug delivery, minimally invasive surgery, and biosensing \cite{wang2021trends}.  Based on their mode of actuation and target application, microrobots range between $\sim$1-1000 $\mu$m in size, and their average speeds vary vastly from $\sim$1 $\mu$m/s to 800 $\mu$m/s, with instantaneous speeds upwards of 800 $\mu$m/s  \cite{li2020micro,jiang2022control}. These properties make microrobots very effective in reaching currently inaccessible areas of the human body but incredibly difficult to visualize and track. Traditionally, microrobots have been studied in aqueous environments (Fig. \ref{fig:motivation}A-B). The growing focus shift in the microrobotic field from system development to biomedical application-oriented implementations necessitates the operation and control of such systems in physiologically relevant environments, and aqueous media do not always represent the conditions and interactions experienced \textit{in vivo}. Furthermore, advancing the current understanding of physical underpinnings of microrobot behaviors \textit{in vivo} requires the study of these systems in physiologically relevant \textit{in vitro}, or \textit{ex vivo}, or \textit{in vivo} environments \cite{soto2020medical}. 

The fast speeds of microrobotic systems (\textit{i.e.}, 10s-100s of body lengths per second) necessitate high frame rate image acquisition, primarily attainable using bright-field imaging. The resulting grayscale images (Fig. \ref{fig:motivation}B) have significantly lower contrast than fluorescent images (Fig. \ref{fig:motivation}A), usually used for automated localization and tracking. Also, microrobots ($\sim$1 $\mu$m in size) often swim in the three-dimensional (3D) space, which translates to intermittent movement of the objects of interest moving in and out of the focal plane, adding another layer of complexity. Moreover, self-propelled micro-motors (\textit{e.g.}, bacteria or catalytic motors) exhibit random walk, making it difficult to track them consistently in every frame. Operation in physiologically relevant environments with dense backgrounds and feature sizes and shapes comparable to those of microrobots (Fig. \ref{fig:motivation}C) further exacerbated this problem. There has been significant progress in real-time tracking of microrobots using clinical imaging modalities that produce high-contrast images \cite{aziz2019real, botros2023usmicromagset,tiryaki2022deep,pane2021real,aziz2020medical,bae2020three,wang2020ultrasound}; however, automated tracking of micro/nano-scale objects in bright-field image sequences within dense backgrounds remains a largely unsolved challenge.   

% The visual ambiguity between the dense background and microscale objects makes manual tracking extremely tedious and time-consuming. Nonetheless, there are a few commonly used techniques for manual or semi-automated tracking of microscale objects. 
A few commonly used techniques exist for manual or semi-automated tracking of microscale objects. Each technique has different features and capabilities, such as particle identification and tracking or morphology analysis.
%The semi-automated methods require user inputs, such as filters or thresholds and manual intervention to prune false positive data.
The most widely used tool for manual tracking is ImageJ (Fiji), an open-source software for processing and analysis of scientific images  \cite{Schneider2012-hb}. Fiji is equipped with several plugins for manual and semi-automated tracking, including Trackmate \cite{TINEVEZ201780,ershov2021bringing}, MtrackJ \cite{MEIJERING2012183}, CellProfiler \cite{mcquin2018cellprofiler,Stirling2021-vf}, MosaicSuite - Particle Tracker  \cite{SBALZARINI2005182}, FARSIGHT \cite{bjornsson2008associative}, BioImageXD \cite{kankaanpaa2012bioimagexd}, and  Icy \cite{de2012icy}. 
% CellProfiler4.0 \cite{Stirling2021-vf} is another open-source software package for image analysis in biology. It includes a module for tracking cells and other objects in microscopy images and videos and allows for the creation of custom analysis pipelines for specific applications. 
The semi-automated methods require user inputs, such as filters or thresholds and manual intervention to prune false positive data. None of these methods were developed for object detection in dense backgrounds, where the visual ambiguity between the dense background and microscale objects makes tracking error-prone, tedious, and time-consuming. 
Apart from these tools, there exist several other methods for particle detection and tracking in 2D and 3D spaces \cite{balomenos2017image,wang2010image, klein2012tlm, CORNWELL2020100440,rust2011single,feng2011multiple,van2008dissecting,stylianidou2016supersegger,gardini20153d}.  However, these methods primarily rely on the high contrast between the object of interest and the background that is unique to fluorescent images. The longer exposure time of fluorescence imaging ($\sim$ 100 ms compared to $\sim$ 10 ms for bright-field imaging) results in a reduced acquisition rate and loss of important temporal data, making it unfeasible for many microrobotic applications. 
% Some methods have explored tracking using brightfield images. For example, Tumblescore \cite{tumblescore} has explored tracking in brightfield images in aqueous medium.  It has the capability of removing noisy data that is unusable for analysis or results in fake points. However, it was primarily designed for less than 10 FPS, which misses the crucial information regarding motility and bacteria behavior. 
Other models have also been developed for tracking objects in aqueous environments using bright-field images. These models consider changes in cells' appearance and overlaps during colony proliferation in time-lapse videos \cite{vallotton2017diatrack,spahn2022deepbacs}. They can also track single cells in aqueous media in bright-field but are still not fully automated and require manual interventions \cite{schwanbeck2020ysmr,xie2008automatic,gardini20153d, mehta2016dissection,doi:10.1073/pnas.1804420115,8481231,tumblescore}. To the authors' best knowledge, automated tracking of micro-motors in dense environments, with features and dimensions similar to those of objects of interest,has not yet been realized. 

In this work, we report the development of a deep learning-based approach for the detection and tracking of microrobots in dense environments. Existing deep learning-based multi-object tracking works have primarily focused on detecting and tracking objects that are easily distinguished from the background (\textit{e.g.}, pedestrians and cars) \cite{milan2016mot16}.  In this work, we present a Motion Enhanced Multi-level Tracker (MEMTrack) for tracking microscale objects in dense environments, where the object of interest is almost indistinguishable from the background features  (Fig. \ref{fig:architecture}C ). We train and test MEMTrack using bacterial micro-biomotors, one of the most commonly used biomotors in biohybrid microrobotic systems  \cite{webster2022biohybrid}. To demonstrate the wide utility of the MEMTrack, we demonstrate its performance in collagen, the most abundant extracellular matrix (ECM) protein in the body, with feature sizes and shapes comparable to those of bacteria (Fig. \ref{fig:motivation}C) and in aqueous media (Fig. \ref{fig:motivation}B). Our results demonstrate that our pipeline can accurately predict and track both visually identifiable and hard-to-detect bacteria. Our proposed pipeline represents a significant contribution to microrobot image analysis in dense environments using computer vision. Moreover, it opens the potential of applying deep learning-based methods for vision-based control of microrobots in dense and low-contrast settings for various applications, including disease diagnosis and treatment.

\section{Methods}
%This section presents the computational methods used for our pipeline of detection and tracking, reviewing state-of-the-art techniques in computer vision as well. We propose MEMTrack, utilizing motion-enhanced feature representations to enhance precision and recall in bacteria detection, particularly in challenging environments. We further discuss the experimental setup used for our research. 

% Begin with computational methods (2.1). Briefly (2-3 sentences) describe the content of the section (key parts of the model) to orient the reader. Then, transition to the description of each model module. For each module of the model, begin with the priors work (a short paragraph), their limitation, then transition to describing our implementation. I suggest we limit how much background information we include here.
% Next is the experimental method (2.2). Reiterate that we use bacteria as the motile part of bacteria-based biohybrid systems and collagen as a tissue surrogate. Describe the experimental work in sufficient detail to ensure reproducibility. Describe the tracking method with the justification of the minimum tracking duration based on the randomization time.
\label{sec:proposed_method}

\subsection{Motion Enhanced Multi-level Tracker (MEMTrack)}
\begin{figure*}[ht]
\centering
  \includegraphics[width=1.0\linewidth]{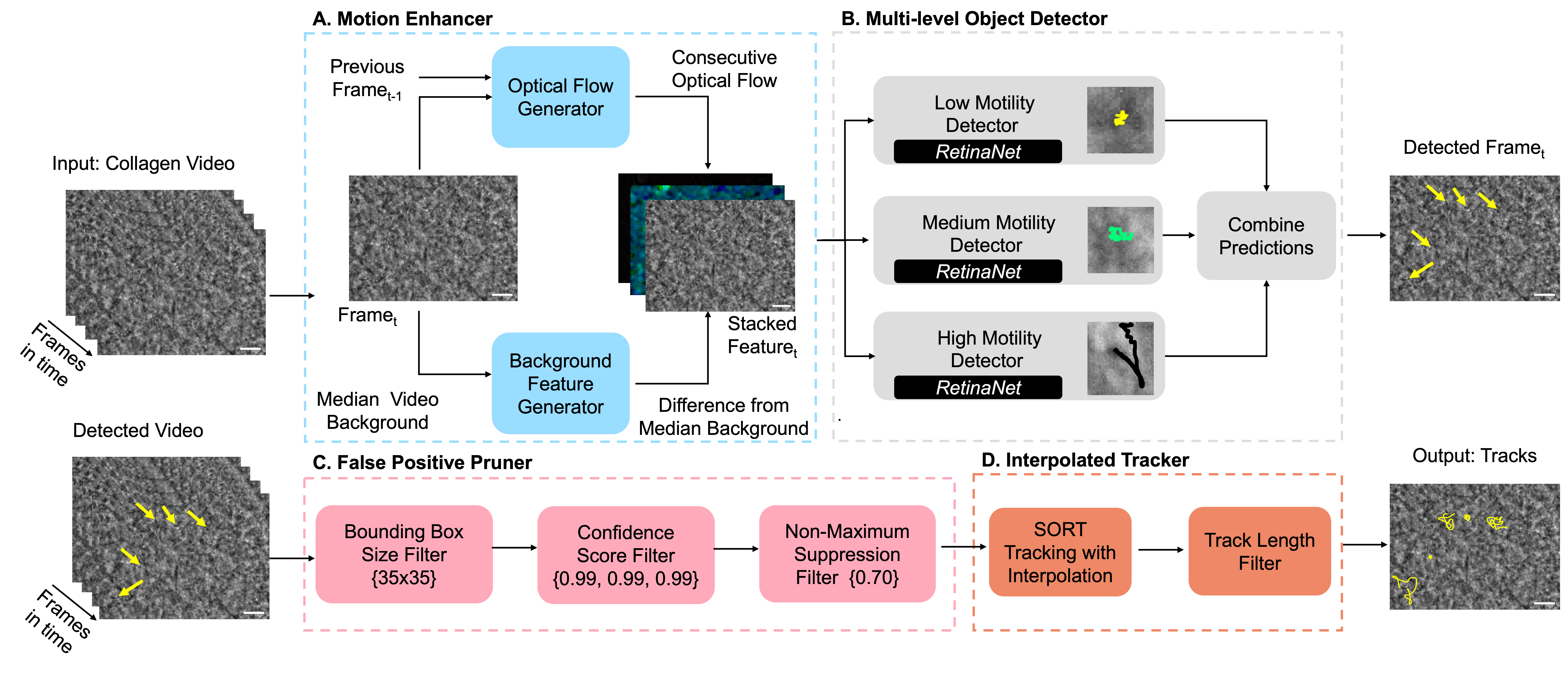}
  \caption{Overview of our proposed pipeline for bacteria tracking consists of four modules: (A) Motion Enhancer, which adds motion features to the input frames, (B) Multi-level Object Detector, which detects objects of varying motility levels using the RetinaNet model, (C) False Positive Pruner, which filters the predicted object occurrences to reduce false positives, and (D) Interpolated Tracker, which tracks objects over time using the SORT algorithm with interpolation.}
  \label{fig:architecture}
\end{figure*}

The proposed pipeline for MEMTrack is shown in Fig. \ref{fig:architecture}. MEMTrack consists of four modules \textemdash  Motion Enhancer, Multi-level Object Detector, False Positive Pruner, and Interpolated Tracker. Before describing each of the modules, we define the notations used throughout the paper. We define the input video with $T$ frames as $I^{1..T} = [I^1, I^2, ..., I^T]$, where $I^t \in \mathbb{R}^{C \times H \times W}$ denotes the $t$-th frame and $C$, $H$ and $W$ are the number of channels, height and width of the image, respectively. We utilized a tracking-by-detection approach, wherein tracking is done on top of predictions from the detection module.

%%%%%%%Architecture Figure so it reflects on the next page%%%%%%%%%%

\subsubsection{Motion Enhancer Module}
% \par \noindent \textbf{Motion Enhancement:} 
% This module seeks to address the limitations of the tracking-by-detection approach when it comes to tracking bacteria. 
Object detectors designed for single-frame detection overlook the object's position in preceding or subsequent frames, which limits their tracking effectiveness. Incorporating the concept of motion into the object detection model is crucial to enable accurate detection of microscale objects in dense environments with intermixing backgrounds and foregrounds (Fig. \ref{fig:motivation}C). Therefore, we implemented feature engineering techniques to capture two types of motion features, optical flow features and median deviation features, which are then augmented or stacked with the image features to enhance detection accuracy, as shown in Fig. \ref{fig:architecture}A and described below. 

\textit{Optical Flow Features.}
Optical flow \cite{article} is a technique that estimates the motion of objects in an image sequence by analyzing the changes in pixel intensities between consecutive frames. We used the Lucas-Kanade \cite{inproceedings} method for optical flow computation, which can be expressed mathematically as:
%which solves an over-determined system of equations using least-squares estimation. 
% \vspace{-1ex}
\begin{equation}
% \vspace{-1ex}
\partial I_x u + \partial I_y v + \partial I_t = 0,
\label{eq:optical-flow}
\end{equation}

\noindent where $u= \frac{dx}{dt}$ and $v =\frac{dy}{dt}$ represent the $x$ and $y$ components of the optical flow vector for the $t$-th frame, $\partial I_x = \frac{\partial I}{\partial x}$, $\partial I_y = \frac{\partial I}{\partial y}$, and $\partial I_t = \frac{\partial I}{\partial t}$ are the image gradients in the $x$, $y$, and time $(t)$ dimensions, respectively. 
% The equation states that the change in pixel intensity over time ($\partial I_t$) is equal to the dot product of the optical flow vector and the image gradient vector, which represents the spatial gradient of the image in the x and y dimensions. 
Solving this equation yields the optical flow vector $O = [u, v]$ for the $t$-th frame in the video. The length of this optical flow vector, which corresponds to the magnitude of motion at each pixel between consecutive frames, is then considered as an additional feature channel along with the grayscale image.

\textit{Median Deviation Features.}
While optical flow captures changes across consecutive frames, we also need features that capture slower motion trends with respect to a static background. To this effect, we used the median deviation as another feature channel, which is the pixel-wise difference between the intensity at a pixel and the median intensity at the pixel across all frames in the video. Note that the median represents the background at every pixel that remains static for the majority of frames. Mathematically, we defined median deviation $\Delta I = |I^t - median(I^{1..T})|$, where $median(I^{1..T})$ is the pixel-wise median for the video across frames $1$ through $T$. 
% The median deviation quantifies how much a pixel's intensity differs from the typical background intensity. A non-zero median deviation suggests that the pixel belongs to a bacterium, indicating deviations from the background.
%The median deviation represents the difference between the pixel intensity at a point and the typical intensity at that point (which we assume is the intensity of the background medium). In other words, it detects if a certain pixel is different from the background, \textit{i.e.} if the pixel belongs to a bacterium, we expect the median deviation to be positive. 

Finally, the input image features $I^t$, concatenated with the optical flow features $O$ and the median deviation $\Delta I$, are used as inputs to the Multi-level Object Detector.

% The median deviation captures the temporal changes in the video. For a video with $n$ frames, each frame being an image of $H\times W\times C$ dimensions, where $C=1$ for bright field agar images, we compute the median of the pixel intensities at each location $(x,y)$ across all frames as computed in \cref{eq:pixel-median}. We then compute the pixel-wise median frame difference for each frame $t$ as the absolute difference between the frame and the median frame:
% \begin{equation}
% \text{med}(I_{1..n}(x,y)) = \text{median}(I_{1}(x,y), I_{2}(x,y), ..., I_{n}(x,y))
% \label{eq:pixel-median}
% \end{equation}

% To create the final image feature, we concatenate the original image, optical flow feature, and frame difference feature. The resulting feature has dimensions $H \times W \times (3C)$, where $C$ is the number of channels in the original image. The final feature can be represented as:

% \begin{equation}
% \mathbf{F}(x,y) = [\mathbf{I}(x,y), \mathbf{O}(x,y), \mathbf{D}(x,y)]
% \end{equation}

% where $\mathbf{I}(x,y)$ is the original image, $\mathbf{O}(x,y,c)$ is the optical flow feature for channel $c$, $\mathbf{D}(x,y,c)$ is the frame difference feature for channel and $c$.$[,]$ denotes concatenation.
% This concatenated feature representation captures both spatial and temporal information, enabling our model to distinguish very hard-to-detect bacteria from the background and makes the approach robust to different concentrations or complexities of background media. 

\subsubsection{Multi-level Object Detector Module}
% \par \noindent \textbf{Multi-level Bacteria Detection:} 

Object detection is a fundamental task in computer vision and plays a crucial role in various applications such as automation and decision-making. The two main approaches used are two-stage detectors and one-stage detectors. Two-stage detectors such as Region-based Convolutional Neural Network (R-CNN) \cite{7112511} and  Fast R-CNN \cite{Girshick_2015_ICCV} offer high accuracy but suffer from high computational complexity, while one-stage detectors such as You Only Look Once (YOLO) \cite{7780460,bochkovskiy2020yolov4, redmon2017yolo9000, wang2022yolov7} provide faster inference but may struggle with detection of small objects in dense environments. 

% Recent versions of YOLO \cite{DBLP:journals/corr/RedmonF16, DBLP:journals/corr/abs-1804-02767,DBLP:journals/corr/abs-2004-10934, li2022yolov6, wang2022yolov7} like YOLOv7 have tried to address these challenges by introducing dynamic label assignment and model structure parameterization (for a comprehensive discussion see ref\cite{10028728}). To address the limitations of one-stage detectors, RetinaNet \cite{DBLP:journals/corr/abs-1708-02002} introduced the novel "focal loss" function, which focuses on hard examples during training, improving accuracy without sacrificing speed. 

In our pipeline, we used the one-stage deep learning detector RetinaNet\cite{lin2017focal} as the base object detector. RetinaNet introduces the novel “focal loss” function, which assigns higher weights to frequently misclassified objects during training, improving accuracy without sacrificing speed. RetinaNet strikes a balance between accuracy and efficiency, making it suitable for real-time tracking in complex backgrounds, as required for microrobot tracking in dense environments. As is the case with most microrobots, bacterial biohybrid microrobots exhibit various motion patterns and speeds ((Fig. \ref{fig:motivation}D-G, Video S1) with different detection requirements, making a single model training for all the observed behaviors ineffective. We propose a Multi-level Object Detection model, where we train a different detector model for each motility category of low, medium, and high (Fig. \ref{fig:architecture}B). The detection models for each category learn specific features and parameters tailored to their unique characteristics. This approach enhances the accuracy of the object detection system for each category and improves the overall performance of the system, as discussed in Section \ref{sec:results}.

During the training phase, the object detector receives input in the form of annotated microscopy videos with bounding boxes measuring 30$\times$30 pixels centered around bacteria centroids. When performing inference, the object detectors predict coordinates $(x, y)$ for bounding boxes to indicate the presence of objects as well as their width and height. These predictions are accompanied by a confidence score that reflects the model's level of certainty regarding the detection.

\subsubsection{False Positive Pruner Module}
The Multi-level Object Detector module can produce a large number of false positives as well as duplicate predictions from the three different detector models. The False Positive Pruner was implemented to remove the false positives without losing the true positive predictions. To this end, first, we combined all the detections from the three motility models and then pruned the detections based on three exclusionary criteria, as shown in Fig. \ref{fig:architecture}C: (1) Bounding Box Filter for removal of predictions that are greater than a prescribed area threshold, (2) Confidence Score Filter for removal of predictions lower than a prescribed confidence threshold, and (3) Non-Maximum Suppression (NMS) \cite{5539960} Filter for elimination of redundant object detections by selecting the ones with the highest confidence score and discarding the other overlapping ones. NMS operates by evaluating the Intersection over Union (
$\text{IoU = AreaOfIntersection/AreaOfUnion}$) between detected bounding boxes. The threshold on IoU serves as a basis to determine whether detected boxes correspond to the same object in the NMS Filter. Beginning with the most confident detected box, NMS eliminates overlapping boxes with lower confidence scores, resulting in accurate object selection and reduced redundancy of detections. 
%Finally, we applied Non-Maximum Suppression (NMS) \cite{5539960} to eliminate redundant object detections by selecting the ones with the highest confidence score and discarding the other overlapping ones. 
% \begin{equation}
% IoU = \frac{\text{Area of Intersection}}{\text{Area of Union}}
% \label{eq:iou}
% \end{equation}
Section \ref{sec:hyper-params} describes our process for selecting the thresholds for the three filters. 
 
\subsubsection{Interpolated Tracker Module}
% \par \noindent \textbf{Interpolated Tracking:} 

Several approaches have been proposed for object tracking, including correlation filter-based \cite{5539960}, Kalman filter-based \cite{bewley2016simple}, and deep learning-based \cite{valmadre2017end, wojke2017simple} methods. Simple Online Real-time Tracker (SORT) \cite{bewley2016simple} is one of the most widely used tracking algorithms. It uses a combination of Kalman filtering \cite{kalman1961new} and the Hungarian algorithm \cite{https://doi.org/10.1002/nav.3800020109, frank2005kuhn} to assign detected objects to existing tracks. The Kalman filter in SORT works by recursively updating estimates of the current system state (in our case, positions of bacteria) based on the previous configuration of the state and the current measurements (\textit{i.e.}, the detected bacteria positions in the current frame) while also taking into account the uncertainty of those measurements. SORT is simple and efficient and performs well on various tracking tasks, such as pedestrian or vehicle tracking. However, the random motion of bacteria and intermittent missing detections resulting from their 3D motion may limit the performance of this algorithm.

In our proposed method, we modify the SORT algorithm by incorporating interpolations for missed object detections. As shown in Fig. \ref{fig:architecture}D, we applied the SORT algorithm to track the detected bacteria and produce tracklets from independent frame-wise predictions. We interpolated the missing detections by keeping the Kalman filter-based unmatched predictions for a given number of frames termed as the maximum age parameter (determined in Section \ref{sec:hyper-params}), and dropping the predictions post that threshold. The maximum age parameter within the SORT algorithm ensures the persistence of a tracklet for a specified number of frames subsequent to a missed detection event, thereby upholding tracking continuity. If a track is missed further than the maximum age, it is discarded.
Finally, we introduce the Track Length Filter to remove tracks whose length, in terms of the number of frames, does not meet a specified threshold (determined in section \ref{sec:hyper-params}). This filter helps in excluding excessively short tracks that are prone to false positives.

\subsection{Experimental Methods}
In order to evaluate the performance of MEMTrack, we recorded bacteria (\textit{i.e.}, the micromotor in bacteria-based biohybrid microrobots) swimming behavior in collagen, as a tissue surrogate, and in an aqueous environment, as the baseline \cite{traore2014biomanufacturing, zhan2022robust}. 

\subsubsection{Bacteria Culture}
\label{sec:bacteriaculture}
Six engineered strains of \textit{Salmonella} Typhimurium VNP20009cheY${^{+}}$ \cite{broadway2017optimizing} bacteria with different motile behaviors were used. Each strain was grown on a 1.5 \%  lysogeny broth (LB; 1 \% tryptone, 0.5 \% yeast extract, and 1 \% sodium chloride) agar plate overnight at 37 ºC.  For each experiment, a single colony of the desired strain was isolated and used to inoculate 10 mL of LB media in a 125 mL smooth-bottom flask. Bacteria were cultured overnight at 37 °C and 100 RPM before being harvested and resuspended in fresh LB to a final concentration of $\sim1.3\times10^6$ CFU/mL. 

\subsubsection{Swimming Assay in Collagen and in Aqueous Medium}
\label{sec:swimmingassay}
Bacteria motility in collagen was evaluated using an experimental setup similar to the traditional swim plate assay, in which bacteria migrate outward from a central inoculation point due to a combination of chemotaxis and growth. Collagen gel was prepared from a stock solution of collagen type I that was neutralized with 0.25 N NaOH, diluted to 5 mg/mL in LB, and supplemented with 100 $\mu$g/mL ampicillin on ice. The cold collagen solution was pipetted into wells of a room-temperature well plate. The well plate was then immediately placed in a 37 °C incubator for 45 minutes to allow the collagen to gel. A 1 $\mu$L aliquot of bacterial suspension, prepared as described in section \ref{sec:bacteriaculture}, was introduced at the center of each collagen well. All data acquisition was performed using a Zeiss AxioObserver.Z1 inverted microscope equipped with a 40$\times$ objective and an hSM camera (Carl Zeiss AG, Oberkochen, Germany). Phase contrast microscopy images were collected at 60 frames per second (FPS) for 2 minutes at 37 ºC.

We also evaluated the motility of each strain in aqueous medium. For swimming speed analysis, a 100 $\mu$L aliquot of the overnight bacteria culture, described in \ref{sec:bacteriaculture}, was used to inoculate a fresh LB culture. The culture was grown at 37 ºC for $\sim$2.5 hours or until $OD_{600}=1.0$ was reached. Subsequently, the culture was diluted $20\times$ in fresh LB medium. A 10 $\mu$L aliquot of the bacterial suspension was placed between two No. 1.5 coverslips separated by a thin ring of vacuum grease. Timelapse imaging was performed as described above at 60 FPS for 10 seconds. For MEMTrack comparison with baseline models, the aqueous swimming assays were performed according to our previously developed methods \cite{zhan2022robust}.

\subsubsection{Bacteria Tracking and Annotation}
\label{sec:annotation}
In order to generate training, validation, and test datasets for MEMTrack, the microscopy videos acquired in collagen and in aqueous media were imported into ImageJ \cite{Schneider2012-hb} software. MTrackJ plugin \cite{MEIJERING2012183} was used to label all bacteria in each frame of each video manually, and their $x$ and $y$ coordinates were recorded. 

\subsubsection{Determining Track Length and Motion Characteristics}
\label{sec:ExpParameters}
In order to determine the tracking period threshold for capturing bacteria random walk in collagen, we first tracked the bacteria for 150 frames and evaluated the diffusivity of each bacterium according to 

\begin{equation}
D(\tau) = \frac{\sum_{t=0}^{2.5} \left((x(t + \tau) - x_0(t))^2 + (y(t + \tau) - y_0(t))^2\right)}{4\tau}
\label{eq:diffusivity}
\end{equation}
where $\tau=0.016$ s is the lag time between consecutive frames, and $x_{0}$ and $y_{0}$ represent the initial position. 

As shown in Fig. \ref{fig:s1}, the diffusivity values plateaued at or before $1$ s, indicating that a track length of $1$ s (60 frames) in collagen is sufficient for capturing the bacteria random walk. Next, we used peak diffusivity values to divide bacteria into four categories based on their motility patterns and the associated diffusivity values\textemdash No motility ($D\leq$ 0.075 \si{\micro\meter\squared\per\second}), low motility ($D\leq$ 0.25 \si{\micro\meter\squared\per\second}), medium motility ($D\leq$ 1 \si{\micro\meter\squared\per\second}), and high motility ($D> 1$ \si{\micro\meter\squared\per\second}).

Based on the well-known motile behavior of bacteria in aqueous environments \cite{berg1993random}, we selected $0.5$ s (30 frames) for testing MEMTrack's performance on aqueous media datasets. 

% \begin{figure}
% \centering
%       \includegraphics[width=1.0\linewidth]{Figures/Picture1.png}
%     \caption{This plot shows that the diffusivity values plateau before 2.5 seconds.}
%     \label{fig:plataeu diff}
% \end{figure}

\subsection{Training, Validation, and Testing Datasets}
\label{sec:model-training} 

MEMTrack was trained and validated using the experimental datasets in collagen, while its performance was evaluated using both the collagen and the aqueous medium datasets. A subset of the collagen data, termed the training set, was used for learning the model parameters (\textit{i.e.}., weights and biases of the deep learning models) using gradient descent algorithms. Simultaneously, another subset of the collagen data termed the validation set, was used during training for observing the performance of the model on data outside the training set and determining the configuration of hyper-parameters (\textit{i.e.}, parameters of the model that are not directly learned using gradient descent such as the filter thresholds of the False Positive Pruner module) that yields best validation performance. Finally, a third subset of the collagen data termed the test set, which has no overlap with the other two sets, was used to report the model performance on “unseen” data. We used the entirety of the aqueous medium data for testing the collagen-trained model in that medium.
To ensure consistency and provide comprehensive training data for different motion types and background scenarios across the six bacterial strains in the collagen, we allocated two videos per strain for training (a total of 12) and reserved one video per strain for validation (a total of 6). The collagen test set consisted of 16 videos representing all bacterial strains. The aqueous medium data set contained 5 videos that were all used for testing.

\subsection{Evaluation Metrics}
We used precision (equation \ref{eq:precision}) and recall (equation \ref{eq:recall}) metrics to evaluate the performance of MEMTrack quantitatively. Precision is the fraction of the true positive (TP) predictions over the sum of all predictions (TPs and false positives (FP)). It is a measure of how precise or confident the model is in tracking real objects (\textit{e.g.}, bacteria). 

\vspace{-1ex}
\begin{equation}
% \vspace{-1ex}
Precision =\frac{TP}{TP +FP}
\label{eq:precision}
\end{equation}

Recall, on the other hand, signifies the fraction of ground truth (GT) or actual bacteria that the model is able to recover. It is defined as the fraction of the TP over the GT (TP and false negatives (FN)).

\vspace{-1ex}
\begin{equation}
% \vspace{-1ex}
Recall =\frac{TP}{TP +FN}
\label{eq:recall}
\end{equation}

Furthermore, we use the F1 score \cite{tan2018introduction} (equation \ref{eq:f1}), which is the harmonic mean of precision and recall, as another evaluation metric to select hyper-parameters in our model architecture.  

\vspace{-1ex}
\begin{equation}
% \vspace{-1ex}
\textit{F1 Score} = 2\frac{Precsion * Recall}{Precision + Recall}
\label{eq:f1}
\end{equation}

\subsection{Determining the Model Hyper-Parameters}
\label{sec:hyper-params}

Model hyper-parameters include the configurational parameters of a machine learning model that are not trained directly using gradient descent but rather need to be set manually before the training begins. The hyper-parameters of the MEMTrack platform are: Area threshold (Bounding Box Filter), Confidence thresholds (Confidence Score Filter), IoU threshold (NMS Filter), Maximum Age parameter (Tracking interpolation), and Track Length threshold (Track Length Filter). 

The Area threshold in the Bounding Box Filter was set to a value of $35\times35$ pixels. Consequently, any detected object characterized by a bounding box area exceeding $35\times35$ pixels was omitted from our predictions. Note that this threshold exceeds the training bounding box size criterion of $30\times30$ pixels to be tolerant of predictions with larger bounding box size during inference compared to training.

\begin{figure*}
\centering
      \includegraphics[width=1.0\linewidth]{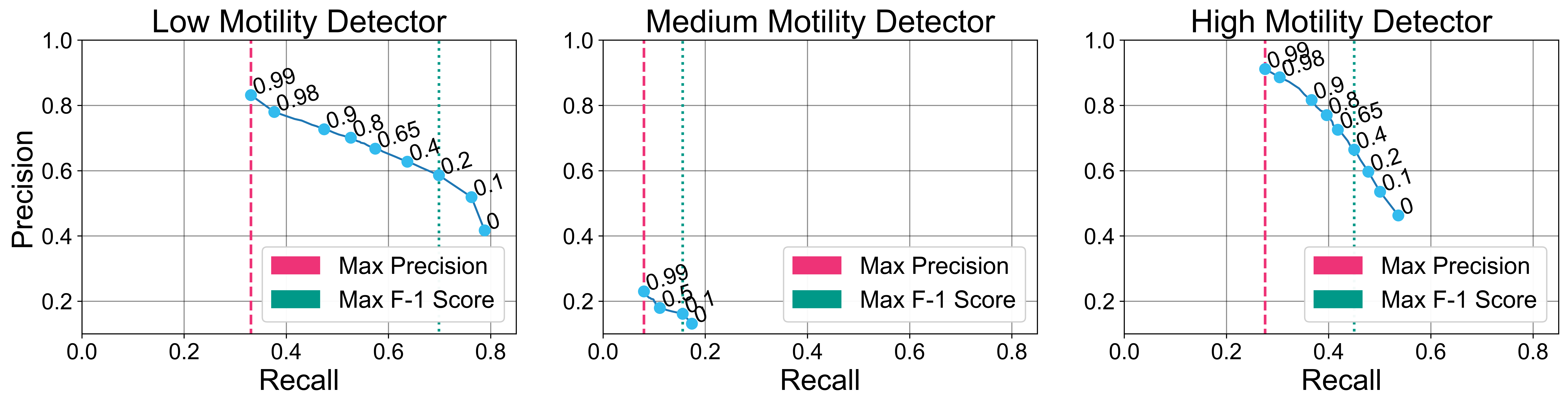}
    \caption{Precision and recall of the three motility detectors (labeled over each plot) for varying confidence score thresholds. Confidence score thresholds of each model based on the maximum precision and maximum F-1 Score criteria are shown as a vertical dashed line and a vertical dotted line, respectively.}
    \vspace{-2.5em}
    \label{fig:collagen_prec_recall_curves}
\end{figure*}
 \vspace{-1.5em}

We experimented with two possible criteria for selecting the confidence score thresholds of the three-level object detector using the validation set\textemdash maximum precision criterion and maximum F1 score criterion. When the maximum precision criterion is used, the three confidence score thresholds (one for each motility level) are set to maximize the precision on the validation set for each motility level. This criterion ensures that the number of false positives is minimized in our predictions across all motility levels. However, by solely maximizing the precision, we cannot control for the number of false negatives and hence may obtain lower recall values. On the other hand, the maximum F1 score criterion ensures that the harmonic mean of precision and recall values are high for all three motility models, thus making a trade-off between reducing false positives and improving recall of the GT bacteria. The choice of maximum precision versus maximum F-1 score criteria depends on the user’s priorities. While the maximum F-1 score criterion would yield a higher recall than the maximum precision criterion, it would do so at the cost of a lower precision value (or, equivalently, more false positive tracks), requiring additional manual intervention to remove the false positive tracks.
% The choice of criteria will impact the Confidence Score Filter threshold  (Figure \ref{fig:architecture}C). We determined the threshold values using the models trained on collagen. 

Fig. \ref{fig:collagen_prec_recall_curves} shows the precision-recall curves on the validation set of the three motility levels for the Multi-level Object Detector, where the threshold settings for the two criteria are indicated as vertical lines. 
For the remainder of this paper, all results from MEMTrack model were obtained using the max precision criterion unless otherwise indicated, since our focus was to obtain predictions of bacteria tracks with high precision without requiring additional manual post-processing steps to remove false positive tracks.  

For the NMS Filter, we used an IoU threshold of 0.7, meaning that if two detected boxes overlap more than 70\%, only the one with the highest confidence score is retained. Note that the threshold is greater than 0.5 since we wanted to retain GT objects that appear in close vicinity (with overlap between 0.5-0.7) as separate detections. 

% This equilibrium ensures the preservation of accurate predictions for true objects in close proximity, while also removing overlapping detections from multiple models. This enhances object selection, leading to improved results and reduced redundancy.

The maximum Age threshold for the SORT Tracking was selected to be 25 frames, indicating that interpolation for absent detections is conducted for up to 25 missed frames; subsequently, tracks beyond this threshold are terminated and discarded. Note that the IoU threshold and Maximum Age threshold settings were manually selected from a pool of other candidate settings to yield the best performance (precision and recall) on the validation set. However, we did not exhaustively search for the best setting of these thresholds using a comprehensive grid search, which we believe could further improve our results.

For the Track Length Filter, we used a threshold of 60 minimum frames to capture bacteria random walk in collagen, as described in  \ref{sec:ExpParameters}. Given the significantly shorter randomization time of bacteria in aqueous environments \cite{leaman2020data}, we used a  minimum track length threshold of 30 frames.  Figure \ref{fig:s2} shows how varying the threshold for minimum track length impacts the precision and recall for our method and the baselines. Increasing this threshold will increase the precision while losing out on many bacteria that have shorter tracks due to them being missed in the detection phase, potentially due to very low-frequency motion. 

\begin{figure*}[b]
\centering
    \includegraphics[width=1.0\linewidth]{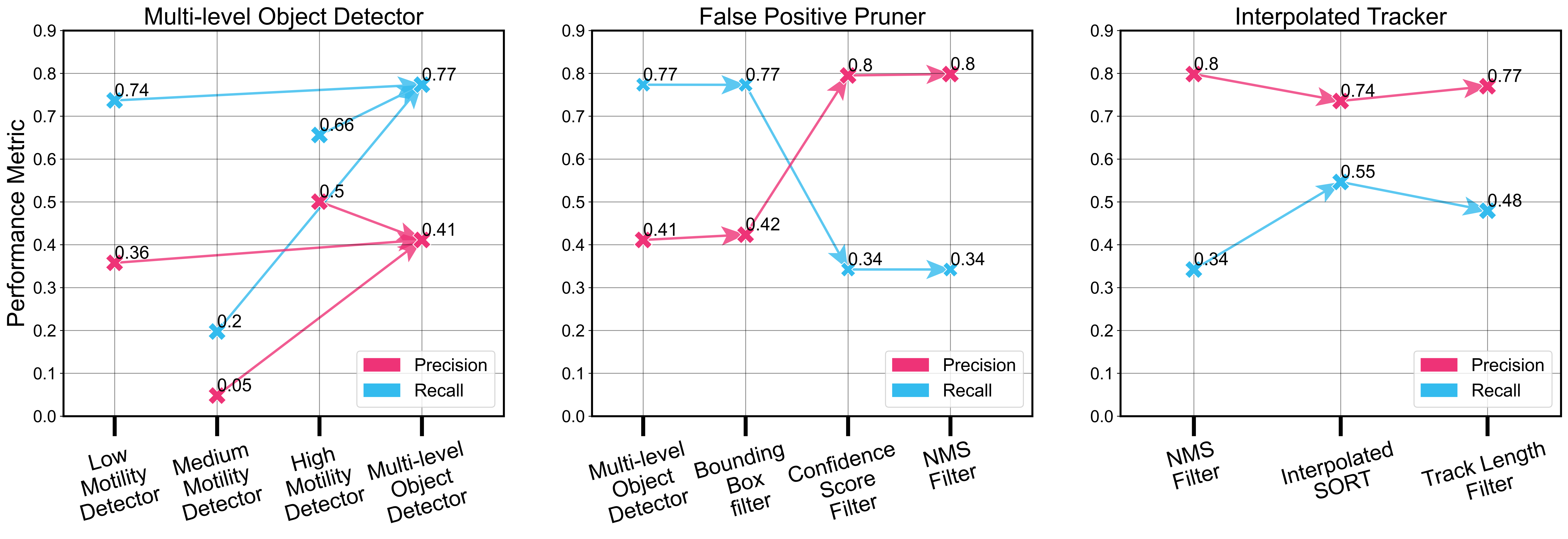}
      \caption{Precision and recall values at each step of the MEMTrack pipeline for the collagen test data.}
      \label{fig:collagen_prec_recall_arrow}
\end{figure*}

\section{Results and Discussion}
% Evaluation of the model performance. In subsection 1 describe the metrics, subsection 2 evaluation in collagen, subsection 3, evaluation in liquid medium.
% Evaluation of human and ML accuracy. In subsection one, we discuss the error analysis (bhas vs himself, model vs himself, model vs bhas). In subsection 2, we discuss speed comparison.
\label{sec:results}

\subsection{Evaluation of Detection Bias}
We first inquired if MEMTrack is able to detect and track bacteria from each of the four motility sub-populations equally well. Such performance capability is crucial to the accurate representation of population-scale behavior. To evaluate the presence of any biases, we segregated the GT bacteria and the TP tracked bacteria into 4 sub-populations based on their motility (refer to Section \ref{sec:ExpParameters}). Table \ref{tab:Motility subpopulation} shows the number of GT and TP bacteria in each of the four categories. The \% Detected column shows the fraction of the GT bacteria outputted from the MEMTrack. It can be seen that MEMTrack has a comparable detection rate for nearly all categories. The lower detection rate for the medium motility group may be attributed to the small number of bacteria in this sub-population, which leads to significant fluctuation in the calculation of  \% Detected.  

\begin{table}[h]
\setlength\tabcolsep{3pt} % let LaTeX figure out the amount of intercolumn whitespace
\fontsize{8.1pt}{9}\selectfont
\centering
\small{
  \centering
  \caption{Fraction of bacteria detected by MEMTrack as a function of motility sub-population}
  \vspace{1ex}
  \label{tab:Motility subpopulation}
  \begin{tabular}{@{}lccc@{}}
    \toprule
    Sub-population & Ground Truth (GT) & True Positive (TP) & \% Detected \\
    \midrule
    No motility      & 17 & 8 & 47 \\
    Low motility      & 140 & 64 & 46 \\
    Medium motility  & 16  & 5  & 31 \\
    High motility & 51 & 25  & 49 \\
    \bottomrule
  \end{tabular}
}
\end{table}

\begin{figure*}
\centering
  \includegraphics[width=0.85\linewidth]{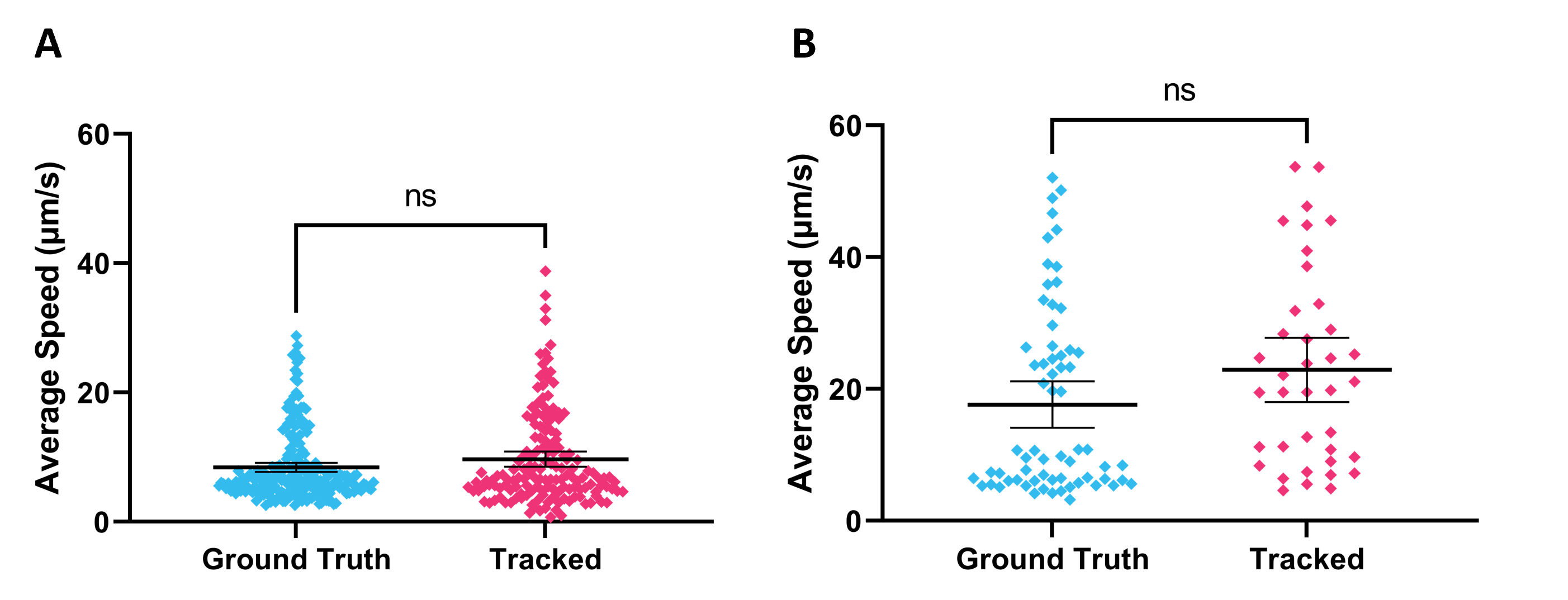}
  \caption{(A) The average speed of the GT bacteria tracked by a human (blue) and the trajectories produced by MEMTrack (pink), irrespective of their match to the GT in collagen (N = 224 for GT, N = 153 for Tracked). (B) The average speed of the GT bacteria tracked by a human (blue) and the trajectories produced by MEMTrack (pink), irrespective of their match to the GT in liquid medium (N = 65 for GT, N = 38 for Tracked).}
  \label{fig:speed comaprison}
\end{figure*}

\subsection{MEMTrack Performance in Collagen}
We next evaluated the MEMTrack performance in the detection and tracking of bacteria in collagen. (Figure \ref{fig:collagen_prec_recall_arrow}, Video S2) shows the precision and recall values attained at each step of the pipeline for the test datasets in collagen. The precision significantly increases after each of the filters of the False Positive Pruner module with a trade-off in the recall. The Interpolated Tracker module further boosts the recall, although at the cost of adding some FPs or a slight reduction in precision. Informed by the physics of bacterial motion in collagen (Section \ref{sec:ExpParameters}), the Track Length Filter removes tracks that are shorter than 60 frames and achieves a reasonable balance between precision and recall. 

To determine MEMTracks ability to accurately describe the average speed of the bacterial population, we compared the average speed of the GT bacteria 8.4$\pm$0.4 $\mu$m/s with the tracked bacteria 9.7$\pm$0.6 $\mu$m/s, as shown in Fig. \ref{fig:speed comaprison}A. A t-test analysis did not indicate any statistically significant difference between the two groups; thus, MEMTrack is able to successfully track bacteria and ascertain population-scale motility speed in the dense collagen environment. 

\begin{table}[b]
\setlength\tabcolsep{3pt} % let LaTeX figure out amount of intercolumn whitespace
\fontsize{8.1pt}{9}\selectfont
\centering
\small{
  \centering
  \caption{Comparison of MEMTrack performance in collagen against two benchmark models.}
  \vspace{1ex}
  \label{tab:collagen-table}
   \begin{tabular}{@{}lccccc@{}}
    \toprule
    \multirow{2}{*}{Method} &  \multicolumn{2}{c}{Tracking} & \multicolumn{2}{c}{Post Track Length Filter (60)} \\
    \cmidrule(lr){2-3} \cmidrule(lr){4-5}
     & Precision & Recall & Precision & Recall \\
    \midrule
    Trackmate  & 0.43 & 0.20 & 0.45 & 0.17\\
    MosaicSuite  & 0.03 & 0.94 & 0.07 & 0.71\\
    MEMTrack (this work)  & \textbf{0.74} & \textbf{0.55} & \textbf{0.77} & \textbf{0.48} \\
    \bottomrule
  \end{tabular}
  }
\end{table}

\subsection{MEMTrack Performance in Aqueous Environment}
Next, we set out to determine the applicability of our model, trained on collagen data, to other media. We evaluated the performance of MEMTrack on bacteria motility data that was collected in an aqueous medium (Video S3).

To determine MEMTracks ability to accurately describe the average speed of the bacterial population, we compared the average speed of the GT bacteria, 17.6$\pm$1.7  $\mu$m/s, with that of the tracked bacteria, 22.9$\pm$2.4 $\mu$m/s, as shown in Fig. \ref{fig:speed comaprison}B. A t-test analysis did not indicate any statistically significant difference between the two groups; thus, MEMTrack is able to track bacteria successfully and ascertain their population-scale motility speed in the aqueous environment as well. 

%\subsubsection{Error Analysis }
%The comparison between GT and tracked bacteria shows that there is a variance at an individual bacterium level. This is why we tried to quantify the error while performing the analysis. Table \ref{tab:model-human-comparison-table} shows that the model has a Root Mean Square Error(RMSE) of 1.7 and 2.64 $\mu$m in $x$ and $y$ coordinates, respectively. While the repeat manual tracking of the same bacteria shows an RMSE of 0.59 and 0.66 $\mu$m in X and Y coordinates respectively.

%\begin{table}
%\setlength\tabcolsep{3pt} % let LaTeX figure out amount of intercolumn whitespace
%\fontsize{8.1pt}{9}\selectfont
%\centering
%\small{
%  \centering
%  \caption{Errors in tracking in micrometers}
%  \label{tab:model-human-comparison-table}
%  \begin{tabular}{@{}lccccccc@{}}
%   \toprule
%    \multirow{2}{*}{Comparison Metric} &  \multicolumn{2}{c}{Average Error($\mu$m)} & \multicolumn{2}{c}{RMSE ($\mu$m)}  & \multicolumn{2}{c}{Absolute Error ($\mu$m)} \\
%    \cmidrule(lr){2-3} \cmidrule(lr){4-5}
%    \cmidrule(lr){6-7}
%     & X & Y & X & Y & X & Y \\
%   \midrule
%    Model vs Human  & 0.089 & 0.366 & 0.972 &  0.838 & 0.504 & 0.544\\
%    Human vs Human  & -0.049 & 0.026 & 0.353 & 0.375 & 0.247 & 0.258 \\
%    \bottomrule
%  \end{tabular}
%  }
%\end{table}

\subsection{Comparing MEMTrack with Baseline Models}
Lastly, we aimed to compare our approach with other existing methods as baselines. %Lastly, We wanted to benchmark our approach against other existing methods. 
Due to the lack of existing models for tracking micro/nano-scale objects in dense environments, we opted for the two most commonly used tracking methods: (1) Trackmate \cite{ershov2021bringing}, and (2) MosaicSuite - Particle Tracker  \cite{SBALZARINI2005182}. We chose these models because they are user-friendly, open source, freely available on the widely used platform FIJI \cite{schindelin2012fiji}, and commonly used for micro/nano-scale object tracking. These tools also have the option to use ML and semi-automated tracking features. Trackmate7 \cite{ershov2021bringing} uses a tracking algorithm based on the Linear Assignment Problem (LAP) tracker  \cite{fukai2023laptrack}, which assigns objects to tracks based on their similarity. MosaicSuite - Particle Tracker  \cite{SBALZARINI2005182} presents a feature point tracking algorithm that is self-initializing, discriminates spurious detections, and can handle temporary occlusion as well as particle appearance and disappearance from the image region. 

We used all the predictions from the benchmark models and compared them against the GT data. We then filtered out tracklets that have a minimum length of 60 frames in collagen and 30 frames in liquid. Results for the comparison of the MEMTrack model and the benchmark methods in collagen and aqueous environments are reported in Tables \ref{tab:collagen-table} and \ref{tab:liquid-table}, respectively. 

Considering a Track Length Filter of 60 frames (or 1 sec), MEMTrack achieved a precision of 77\% precision and a recall of 48\% in collagen (Table \ref{tab:collagen-table}). In contrast, Trackmate had significantly lower precision and recall values of 45\% and 17\%, respectively. MosaicSuite showed the highest recall at 71\% but at the cost of a significantly lower precision of 7\% (\textit{i.e.}, only 7\% of tracks are TPs), making its results unusable without extensive manual post-processing for filtering out the FPs. Overall, our approach can provide significantly better precision than the benchmark methods with reasonably high recall.

\begin{table}[t]
\setlength\tabcolsep{3pt} % let LaTeX figure out amount of intercolumn whitespace
\fontsize{8.1pt}{9}\selectfont
\centering
\small{
  \centering
  \caption{Comparison of  MEMTrack performance in liquid media against two benchmark models.}
  \vspace{1ex}
  \label{tab:liquid-table}
   \begin{tabular}{@{}lccccc@{}}
    \toprule
    \multirow{2}{*}{Method} &  \multicolumn{2}{c}{Tracking} & \multicolumn{2}{c}{Post Track Length Filter (30)} \\
    \cmidrule(lr){2-3} \cmidrule(lr){4-5}
     & Precision & Recall & Precision & Recall \\
    \midrule
    Trackmate  & 0.84 & 0.39 & 0.83 & 0.37\\
    MosaicSuite  & 0.27 & 0.94 & 0.39 & 0.86\\
    MEMTrack -Max Precision   & \textbf{0.94} & \textbf{0.20} & \textbf{0.94} & \textbf{0.15} \\
 MEMTrack -Max F-1 Score   & \textbf{0.88} & \textbf{0.40} & \textbf{0.94} & \textbf{0.35} \\
  % MEMTrack -Max Recall   & \textbf{0.72} & \textbf{0.60} & \textbf{0.75} & \textbf{0.51} \\
    \bottomrule
  \end{tabular}
  }
\end{table}

Table \ref{tab:liquid-table} compares the performance of the pre-trained (in collagen) MEMTrack with the same Trackmare and MosaicSuite benchmark models in aqueous media. Consistent with the results in collagen, MEMTrack performed better than both benchmark models with a precision of 94\% and a recall of 15\% on the maximum precision setting and a precision of 94\%, and a recall of 35\% on the maximum F-1 score setting. Trackmate produced somewhat comparable results with a precision of 83\% and a recall of 37\%. MosaicSuite's precision for liquid data was 39\%, which was significantly improved compared to that of collagen data (7\%); nonetheless, its significantly low precision necessitates manual intervention to remove false positives. As expected, the application of the Track Length Filter improved the precision in most cases and decreased the recall (Fig. \ref{fig:s2}). Altogether, these results demonstrate that our model can be used to effectively track micro/nano-scale objects in liquid media without additional training. MEMTrack performance in liquid media can potentially be further improved if we re-train models from scratch on liquid data instead of using the pre-trained Collagen models.

% \subsection{Ablation Studies}

% To further evaluate the effectiveness of our proposed pipeline, we conducted ablation studies on the size of the bounding box used for detecting bacteria. We trained bacteria detection models with varying bounding box sizes based on the average size of bacteria in our dataset, 22 pixels. Our experiments in Table \ref{tab:bbox_size_ablation} show that using the appropriate bounding box size significantly improves the detection accuracy, with larger boxes leading to more false positives and smaller boxes leading to missed detections. Additionally, we performed experiments on applying non-maximum suppression (NMS) on the combined detections to remove redundant detections. Our results, Table \ref{tab:nms}, demonstrate that applying NMS effectively reduces the number of duplicate tracks while maintaining a high recall rate. Overall, these ablation studies highlight the importance of carefully selecting the bounding box size and applying NMS to improve the accuracy and efficiency of our proposed pipeline for tracking bacteria in complex environments.

\section{Conclusion}
% Highlights key contributions of the paper and their impact
% Briefly describe limitations in the context of future work.
In this work, we present MEMTrack, an automated pipeline for detecting and tracking microrobots in dense and low-contrast environments, such as collagen. This is a particularly challenging problem given the lack of visual features distinguishing the foreground objects from the background. Our approach leverages synthetic motion features, the RetinaNet object detection model and a modified SORT tracking algorithm with interpolation to achieve robust results against different background media and with high precision. Our test results demonstrate that the proposed pipeline can robustly substitute the tedious task of manually tracking microrobots for the prediction of population-scale speed values. Moreover, we show the broader applicability of our method by using it to track bacteria in liquid media.
Our pipeline also has limitations that can be improved in future research. For instance, instead of using the SORT tracking algorithm, which only uses the positions of detected bacteria to perform tracking, deep learning-based tracking methods such as DeepSORT \cite{wojke2017simple} can be used in future extensions to take into account the visual features of detected bacteria in every frame to perform tracking. Learning low-frequency motion patterns of bacteria from longer periods can also be explored to enhance accuracy in tracking no and low-motility bacteria categories. Finally, the tracking algorithm can be improved by including physics-based knowledge of bacteria motion in varying environments to guide the Interpolated Tracker module.

% \medskip
% \textbf{Supporting Information} \par %Please delete the Suppporting Information statement if it is not applicable. Please supply Supporting Information in another file. Supporting information should not be provided in .tex format
% Supporting Information is available from the Wiley Online Library or from the author.

% \section*{Conflicts of interest}
% There are no conflicts to declare.

% Acknowledgements
% \medskip
% \textbf{Acknowledgements} \par %delete if not applicable))
\section{Acknowledgements}
The authors acknowledge Behkam Lab member, Ying Zhan, for sharing previously published bacteria aqueous swimming data to test MEMTrack performance in speed predictions. This research was supported in part by NSF grants CBET-2133739 and CBET-1454226 to BB, 4-VA grant to BB, and NSF grant IIS-2107332 to AK. Access to computing resources was provided by the Advanced Research Computing (ARC) Center at Virginia Tech.

% References
% \medskip

% Use the following code if you wish to generate your bibliography with BibTeX;
% replace the string "MSP-template" below with the name(s) of
% the BibTeX data base(s) you want to use.
% The resulting bibliography-output (the content of the .bbl file)
% must be pasted back into this file before submission.
% Please also include your BibTeX data base file(s) in your submission
% so that we can re-run BibTeX if necessary.
%

% \bibliography{ref}

% Table of contents entry should be 50 - 60 words long
% Image should be 55 mm broad and 50 mm high or 110 mm broad and 20 mm high

\bibliography{ref}
\bibliographystyle{iclr2024_conference}

\newpage
\appendix
\section{Supplementary Videos}

All supplementary video are publicly available at this link: (\url{{https://doi.org/10.5281/zenodo.10001477}}).
This repository contains 3 videos as detailed below:

\textbf{Video S1}: Video shows the movement of representative bacterial biomotors from each of the four motility subpopulations in collagen. Video slowed down by 2x 

\textbf{Video S2}: Videos show ground truth annotation (red) by manual tracking and MEMTrack-enabled automated detection and tracking of bacteria in collagen. All scale bars are 20 µm and timestamp indicates s:ms. Video slowed down by 2x 

\textbf{Video S3}: Videos show ground truth annotation (red) by manual tracking and MEMTrack-enabled automated detection and tracking of bacteria in aqueous media. All scale bars are 20 µm and timestamp indicates s:ms. Video slowed down by 2x 

\vspace{5ex}

\section{Supplementary Figures}

\begin{figure}[h]
\centering
    \includegraphics[width=0.5\linewidth]{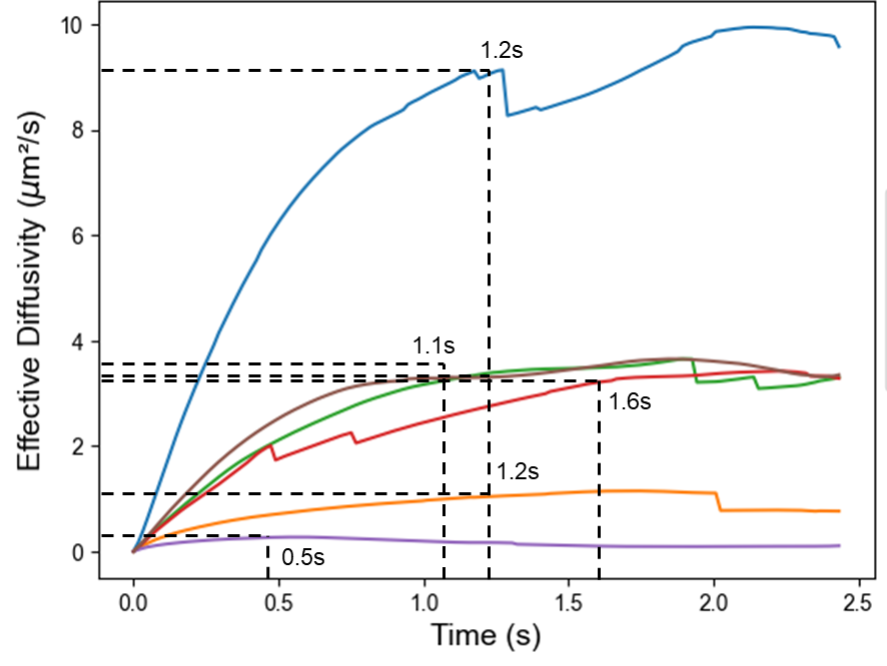}
      \caption{Bacteria diffusivity in collagen plateaus after approximately 1s.}
      \label{fig:s1}
\end{figure}

\begin{figure} [ht]
\centering
    \includegraphics[width=1.0\linewidth]{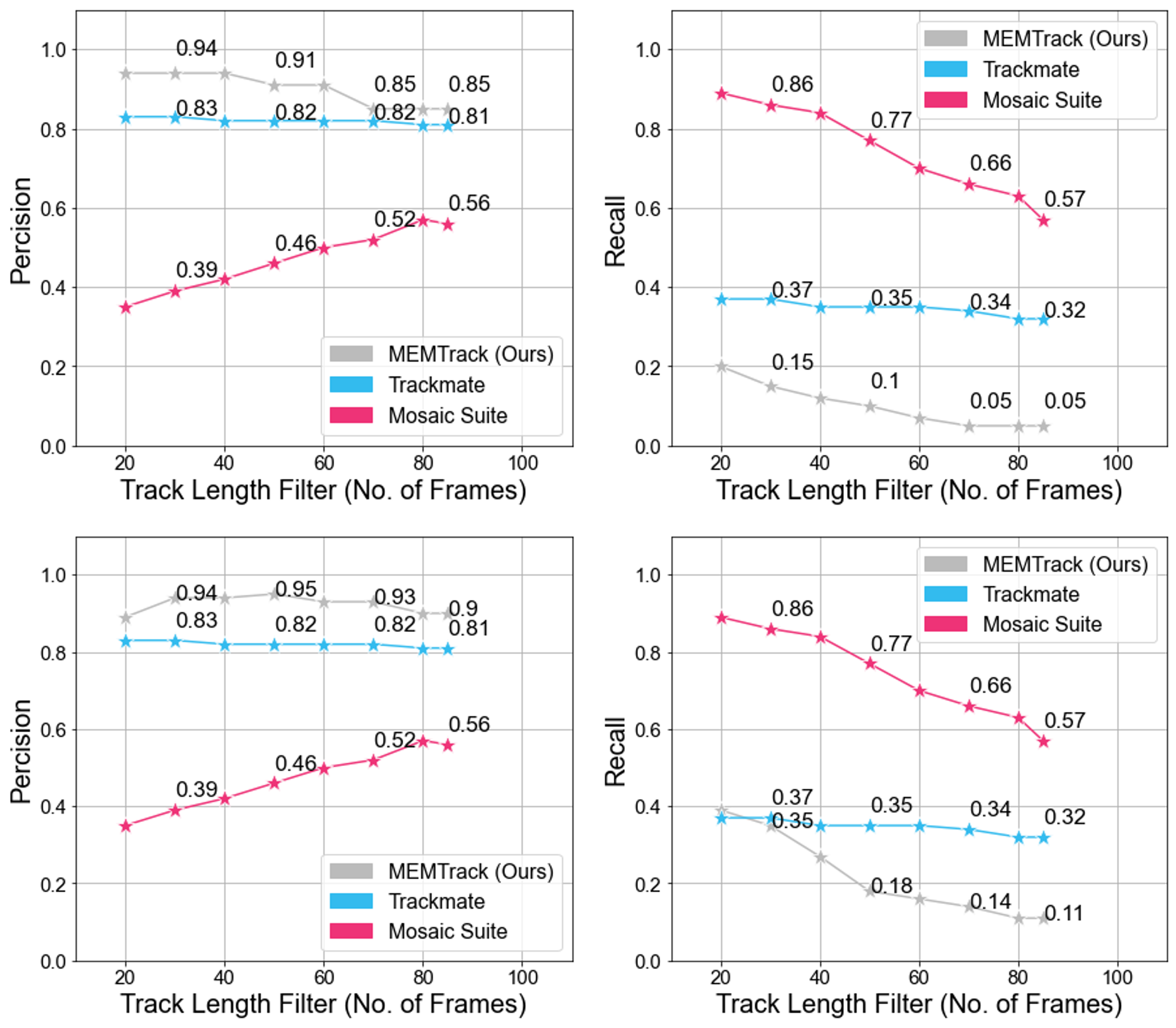}
      \caption{Effect of the Track Length Filter threshold on the precision and recall values for the liquid dataset. The top row shows the results with max-precision criteria, and the bottom row shows the results for max F1 criteria.}
      \label{fig:s2}
\end{figure}

% \begin{figure}[h]
% \centering
%     \includegraphics[width=1.0\linewidth]{SupplementaryFigures/S3.png}
%       \caption{Results on Liquid Data. Ground Truth Bacteria: Red Predicted Bacteria: Yellow}
%       \label{fig:s3}
% \end{figure}

% \begin{figure}[h]
% \centering
%     \includegraphics[width=0.45\linewidth]{SupplementaryFigures/S4.png}
%       \caption{Comparison of the average speed of bacteria in collagen using manually tracked ground truth data and the true positive data from MEMTrack. A t-test statistical analysis shows no significant (ns) different between the two groups. n=150 and n=103 for the ground truth and the true positive categories.}
%       \label{fig:s4}
% \end{figure}

% \begin{figure}[h]
% \centering
%     \includegraphics[width=0.45\linewidth]{SupplementaryFigures/S5.png}
%       \caption{Comparison of the average speed of bacteria in aqueous medium using manually tracked ground truth data and the true positive data from MEMTrack. A t-test statistical analysis shows no significant (ns) different between the two groups. n=40 and n=26 for the ground truth and the true positive categories.}
%       \label{fig:s5}
% \end{figure}
% \section{Appendix}
% You may include other additional sections here.

\end{document}